\documentclass[10pt,twocolumn,letterpaper]{article}

\usepackage{cvpr}              % To produce the CAMERA-READY version
\usepackage{comment} % 追加した
\usepackage{array}
\usepackage{tabularx}
\usepackage{multirow} 

\definecolor{cvprblue}{rgb}{0.21,0.49,0.74}
\usepackage[pagebackref,breaklinks,colorlinks,allcolors=cvprblue]{hyperref}
\usepackage{tcolorbox} % 

\usepackage{soul}
\usepackage{xcolor}
\definecolor{ngreen}{HTML}{4daf4a}
\definecolor{nblue}{HTML}{377eb8}
\definecolor{nred}{HTML}{e41a1c}
\definecolor{oorange}{HTML}{F4C842}

\newcommand{\hlc}[2][yellow]{{%
    \colorlet{foo}{#1}%
    \sethlcolor{foo}\hl{#2}}%
}

\def\paperID{22} % *** Enter the Paper ID here
\def\confName{CVPR}
\def\confYear{2026}

\title{Responses Fall Short of Understanding: Revealing the Gap between Internal Representations and Responses in Visual Document Understanding}

\author{
Haruka Kawasaki \quad Ryota Tanaka \quad Kyosuke Nishida\\
Human Informatics Labs., NTT, Inc.\\
{\tt\small\{haruka.kawasaki, ryota.tanaka, kyosuke.nishida\}@ntt.com}}

\begin{document}
\maketitle
\begin{abstract}
Visual document understanding (VDU) is a challenging task for large vision language models (LVLMs), requiring the integration of visual perception, text recognition, and reasoning over structured layouts. Although recent LVLMs have shown progress on VDU benchmarks, their performance is typically evaluated based on generated responses, which may not necessarily reflect whether the model has actually captured the required information internally. In this paper, we investigate how information required to solve VDU tasks is represented across different layers of LLMs within LVLMs using linear probing. Our study reveals that (1) there is a clear gap between internal representations and generated responses, and (2) information required to solve the task is often encoded more linearly from intermediate layers than from the final layer. Motivated by these findings, we explore fine-tuning strategies that target intermediate layers. Experiments show that fine-tuning intermediate layers improves both linear probing accuracy and response accuracy while narrowing the gap.

\end{abstract}

\newcolumntype{C}{>{\centering\arraybackslash}X}

% ====================================================================
\section{Introduction}
\label{sec:intro}

\begin{figure}[t]
\centering
 \includegraphics[width=\linewidth,clip]{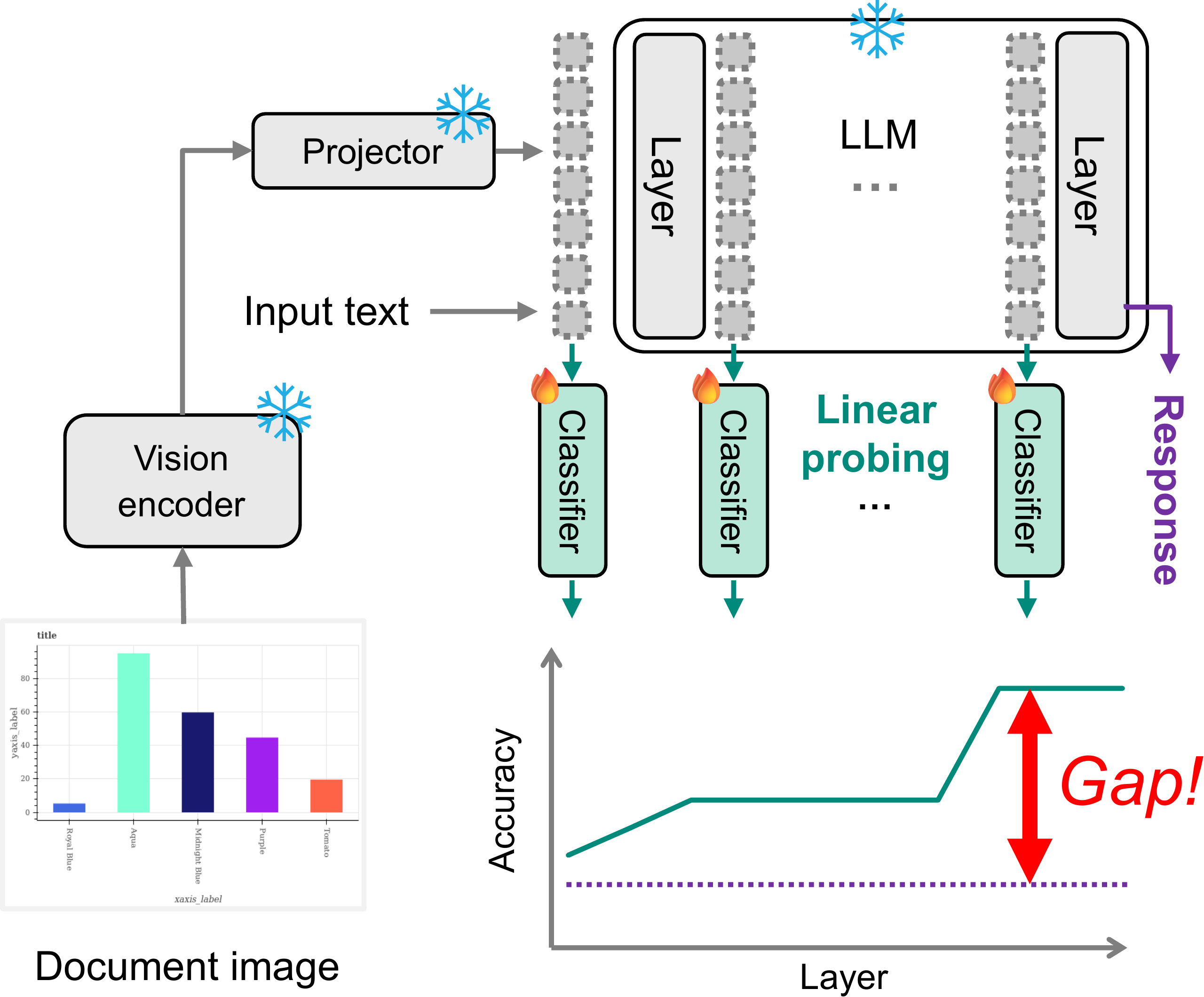}
    \caption{\textbf{Overview of our analysis.}
    We analyze the gap between the internal representations and responses in VDU.
    For the analysis of internal representations, we employ linear probing and construct classifiers at each layer. For the response, we evaluate the accuracy of text responses. }
\label{fig:overview}
\end{figure}

% -------------------------------
Large vision language models (LVLMs) have recently shown remarkable progress across a wide range of multimodal tasks~\cite{chen2024expanding,Qwen2.5-VL}. 
In particular, LVLMs achieve strong performance in visual document understanding (VDU), an important and challenging problem that requires models to integrate visual perception, text recognition, and reasoning over structured document layouts. 

Despite this progress, the evaluation of LVLMs in VDU primarily relies on the correctness of generated responses. 
However, response accuracy alone may not fully reflect whether a model has internally captured the information necessary to answer a question. 
Recent studies~\cite{gekhman2025insideout,chandhok-etal-2025-response} have suggested that internal representations can contain richer information than what is expressed in generated responses. 
In addition, the representation-based evaluation paradigm~\cite{li2026rethinkingllmasajudgerepresentationasajudgesmall} shows that hidden representations of language models can provide reliable signals for evaluation without relying on generation. These studies suggest that models may internally encode useful information that is not reflected in their responses, revealing a gap between internal representations and responses.
Understanding this gap between internal representations and responses is important not only for evaluating performance but also for enhancing the reliability, interpretability, and safety.
Because VDU requires integrating multiple modalities and structured reasoning over layouts, it provides a particularly challenging testbed for analyzing how multimodal information is represented within LVLMs.

In this paper, we investigate how information required to solve VDU tasks is represented across layers of LLMs within LVLMs using linear probing (see Figure~\ref{fig:overview}).
Linear probing provides a simple yet effective way to evaluate whether information is linearly encoded from hidden representations. Our analysis reveals two key findings. 
\textbf{First}, we observe a clear gap between internal representations and responses in VDU tasks: information that can be linearly encoded from hidden representations is not always reflected in the model’s responses. 
\textbf{Second}, we find that the most linearly encoded representations of information required to solve VDU tasks often appear in intermediate layers rather than in the final layer.

Motivated by these findings, we further investigate whether this gap can be reduced by leveraging intermediate layers. 
We explore fine-tuning strategies that target selected intermediate layers of LLMs within LVLMs, guided by insights obtained from our linear probing analysis.
Experimental results show that fine-tuning all the model's layers (all-layer fine-tuning) is insufficient to fully close the gap. On the other hand, fine-tuning intermediate layers efficiently improves both linear probing accuracy and response accuracy, while also reducing the gap between internal representations and responses more than all-layer fine-tuning.

Our main contributions are summarized as follows:
\begin{itemize}
    \item We conduct a comprehensive layer-wise linear probing analysis of LVLMs for VDU tasks.
    \item We demonstrate that a significant gap exists between internal representations and responses in VDU tasks, which cannot be reduced by all-layer fine-tuning.
    \item We show that the information required to solve VDU tasks is often more accessible in intermediate layers than in the final layer.
    \item We demonstrate that fine-tuning selected intermediate layers improves response accuracy and reduces the gap between internal representations and responses.
\end{itemize}

% -------------------------------

\begin{figure*}[t]
\centering
\includegraphics[width=\linewidth,clip]{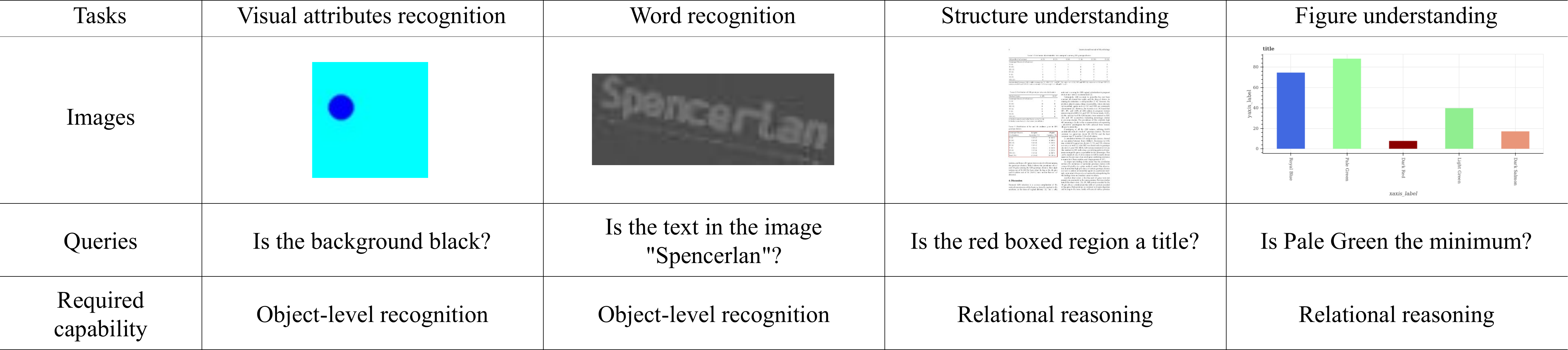} 
    \caption{\textbf{Examples of linear probing tasks.}
    We use four linear probing tasks covering different aspects of VDU.
    These include visual attributes recognition, which targets properties such as color and shape; word recognition, which focuses on identifying spelling differences between the word in the image and the query; structure understanding, which asks about the document component highlighted in the image; and figure understanding, which requires reasoning over graphical elements in charts.}
\label{fig:tasks}
\end{figure*}

% =====================================================================
\section{Related Work}
\label{sec:related_work}

\paragraph{Visual document understanding.}

Understanding documents from visual inputs, known as  VDU, has become an important capability for both research and industrial applications of LVLMs. 
Recent advances have enabled LVLMs to process document images and perform tasks such as document question answering~\cite{hu-etal-2024-mplug,hu-etal-2025-mplug,chen2024far,gao2024mini,chen2024expanding,ye-etal-2023-ureader,wang2024qwen2,Qwen2.5-VL,agrawal2024pixtral}. 
As a result, LVLMs have demonstrated strong performance on several document understanding benchmarks~\cite{VisualMRC2021,SlideVQA2023,van2023document,zhu2022towards,NEURIPS2024_ae0e4328}.
However, compared to natural image understanding, document understanding remains challenging~\cite{fu2024ocrbenchv2improvedbenchmark}.
This is because the information required to solve the question is often distributed across multiple regions and structured layouts. 
Moreover, VDU requires models to integrate visual and textual cues across the document, rather than relying on a single localized signal. 
As a result, accurately assessing whether a model has internally captured the necessary information is challenging.
Most existing studies evaluate LVLMs in VDU based solely on the correctness of generated responses, leaving the internal representations underlying their predictions largely unexplored. 
Understanding how information required to solve the question is internally represented in LVLMs is therefore important for analyzing their reasoning behavior in VDU tasks.
In this work, we conduct a detailed layer-wise analysis of internal representations in LVLMs for VDU tasks using linear probing.

\paragraph{Internal representations analysis using linear probing.}

Linear probing~\cite{alain2016understanding} is one of the most commonly used methods for determining whether the representations produced by a model contain sufficiently informative features. 
This enables a more direct analysis of the relationship between internal representations and model outputs.
Linear probing analysis has been extended to LVLMs~\cite{chandhok-etal-2025-response,huang-etal-2025-vision}. 
Although prior work has compared the final-layer internal representations with responses, a comparison of representations across the entire set of layers remains largely unexplored.
In this work, we conduct a detailed layer-wise linear probing analysis to investigate how information required to solve VDU tasks is represented in LVLMs and investigate the gap between internal representations and generated responses.
In addition, we use findings from linear probing analysis to design fine-tuning strategies targeting intermediate layers. 
Our results demonstrate that insights from representation analysis can improve model performance.

\paragraph{Gap between internal representations and responses.}
Recent work has investigated whether LLMs internally encode factual knowledge and truthfulness signals beyond what is observable in their responses.
Several studies~\cite{azaria2023internal,marksgeometry} demonstrate that truth-related information is accessible in hidden representations. 
However, prior work has shown that internal signals may not consistently align with generated responses, highlighting a gap between internal representations and responses~\cite{liu2023cognitive}.
Building on this gap, a method that estimates entity knowledge directly from internal entity representations was proposed~\cite{gottesman2024estimating}. 
In addition, studies that further analyze and quantify discrepancies between internal and external knowledge are conducted~\cite{orgad2025llms,gekhman2025insideout}. 
It has also been demonstrated that LLM can leverage internal states to estimate answerability and improve factuality prediction~\cite{ni-etal-2025-towards}. 
Collectively, these works suggest that LLMs often encode more information in internal representations than is directly reflected in generated textual responses.
This gap has also been observed in LVLMs~\cite{chandhok-etal-2025-response}.
However, the internal representations of LLMs in LVLMs have not yet been layer-wise examined, and whether the most informative representations arise in the final layer or earlier layers remains largely unexplored in LVLMs.
In addition, an investigation of this phenomenon for VDU is also largely unexplored. 
In this study, we address this gap by conducting a detailed layer-wise analysis of LLMs in LVLMs to determine whether the phenomenon also arises in VDU.

% =====================================================================
\section{Linear Probing Experiment}
\label{sec:lp} 
We perform a detailed \textbf{layer-wise} analysis of the internal representations of LLMs within LVLMs using linear probing.
Specifically, we train binary classifiers on intermediate representations extracted during inference for linear probing tasks.
We evaluate linear probing accuracy to assess the extent to which information required to solve the question is linearly encoded in representations at each layer.
In addition, by comparing linear probing accuracy with the model’s response accuracy, we investigate whether the internal representations are reflected in its responses.

\subsection{Building Linear Classifiers}

\paragraph{Inputs for linear classifiers.}
We construct separate linear classifiers for each LLM layer in LVLMs and for each token type. 
The image is provided to the LVLM before the query text input.
At every layer, we build four classifiers that take as input as follows: (1) \hlc[nblue!30]{image-token}: tokens corresponding to visual embeddings projected from the vision encoder outputs, (2) \hlc[ngreen!30]{text-token}: tokens corresponding to the textual inputs of the question, (3) \hlc[oorange!30]{all-token}: all tokens in the sequence, and (4) \hlc[red!30]{last-token}: the last token of the sequence.

\paragraph{Classifier architecture.}
Let $\{\mathbf{h}_i\}_{i \in \mathcal{T}}$ with $\mathbf{h}_i \in \mathbb{R}^d$ be the hidden states of a given token type at a given LLM layer, where $\mathcal{T}$ is the index set of the token type and $d$ is the hidden dimension.
For (1) \hlc[nblue!30]{image-token}, (2) \hlc[ngreen!30]{text-token}, and (3) \hlc[oorange!30]{all-token}, we compute the mean-pooled vector:
\[
\bar{\mathbf{h}} = \frac{1}{|\mathcal{T}|} \sum_{i \in \mathcal{T}} \mathbf{h}_i,
\]
while for the (4) \hlc[nred!30]{last-token} setting, we directly use the last token's hidden state.
Let $\mathbf{h} \in \mathbb{R}^d$ be the resulting input to the classifier.
The classifier is a single linear layer:
\[
\mathbf{z} = W \mathbf{h} + \mathbf{b}, \qquad \mathbf{z} \in \mathbb{R}^{2},
\]
with trainable parameters $W \in \mathbb{R}^{2 \times d}$ and $\mathbf{b} \in \mathbb{R}^{2}$.
The output logits $\mathbf{z}$ are optimized using the cross-entropy loss between the predictions and the targets, while keeping the LVLM parameters frozen.

\subsection{Linear Probing Tasks}
\label{subsec:data}
We construct four binary classification tasks (see Figure~\ref{fig:tasks}).
To ensure comparable difficulty and to focus on non-trivial cases, we select samples for which Qwen2.5-VL 3B generates incorrect answers, where correctness is determined by whether the generated response contains the ground-truth answer. During dataset construction, 78\% of the samples are filtered out.
Our evaluation dataset requires two capabilities: (1) \textbf{object-level recognition}, which requires the model to identify visual elements such as objects, colors, shapes, and even individual characters within a word in an image, and (2) \textbf{relational reasoning}, which requires the model to interpret structural elements in document images and to reason about relationships between visual attributes in charts and figures. To evaluate these capabilities, we prepared four tasks as follows:

\paragraph{Visual attributes recognition.}
This task is constructed using the easy-VQA dataset~\cite{easy-VQA_github}. 
This consists of images and questions that target visual attributes such as color and shape.
For example, this includes queries such as “\textit{Is the background black?}”.
The dataset consists of 100,000 training samples and 10,000 test samples.

\paragraph{Word recognition.}
This task leverages images from the MJSynth Text Recognition dataset \cite{Jaderberg14c,Jaderberg16}.
We created queries in each image to evaluate word recognition. 
Negative examples are generated from incorrect predictions generated by Qwen2.5‑VL 3B when reading the images.
For example, queries include “\textit{Is the text in the image ‘Spencerlan’?}”, where ‘Spencerlan’ is an incorrect prediction for the correct word ‘Spencerian’.
The dataset consists of 100,000 training samples and 10,000 test samples.

\paragraph{Structure understanding.}
This task uses images and annotations from the PubLayNet \cite{zhong2019publaynet}.
We render red bounding boxes based on the annotations.
The task is to identify the document component contained within the highlighted region.
The document component types are title, text, list, figure, and table.
Queries are attached to each sample, such as “\textit{Is the red boxed region a title?}”.
The dataset consists of 100,000 training samples and 10,000 test samples.

\paragraph{Figure understanding.}
This task is based on the FigureQA dataset \cite{kahou2017figureqa}.
This requires comparative reasoning over elements within a chart.
For example, this includes queries such as “\textit{Is Pale Green the minimum?}”.
The dataset contains 74,913 training samples and 6,111 test samples.

\begin{figure*}[htbp]
  \centering

\begin{minipage}{\textwidth}
\centering  
  \begin{minipage}{0.245\linewidth}
    \centering
    \includegraphics[height=0.96\linewidth,clip]{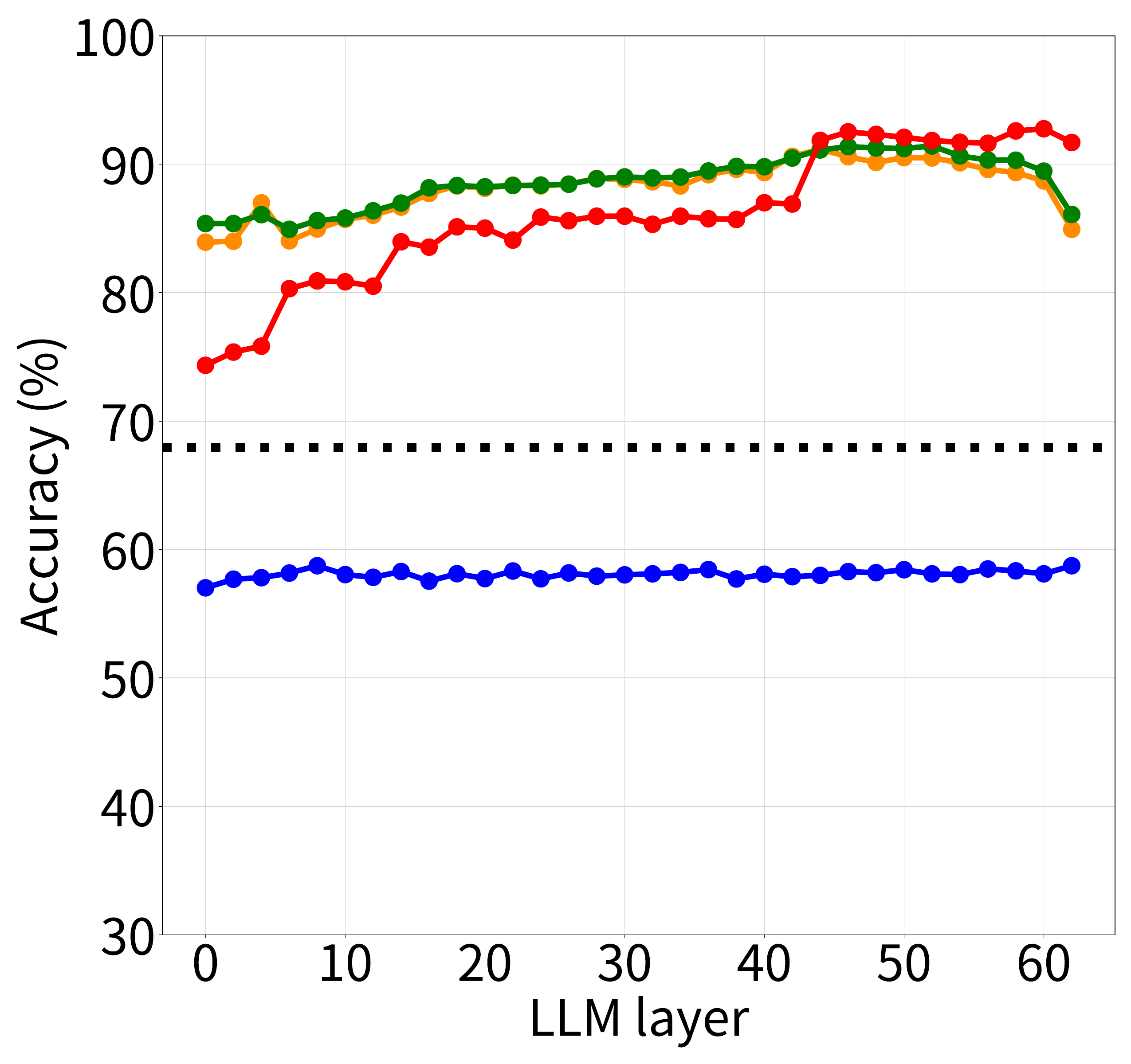}
    {\small \hspace*{1.8em}Visual attributes recognition}
  \end{minipage}
  \hfill
  \begin{minipage}{0.245\linewidth}
    \centering
    \includegraphics[height=0.96\linewidth,clip]{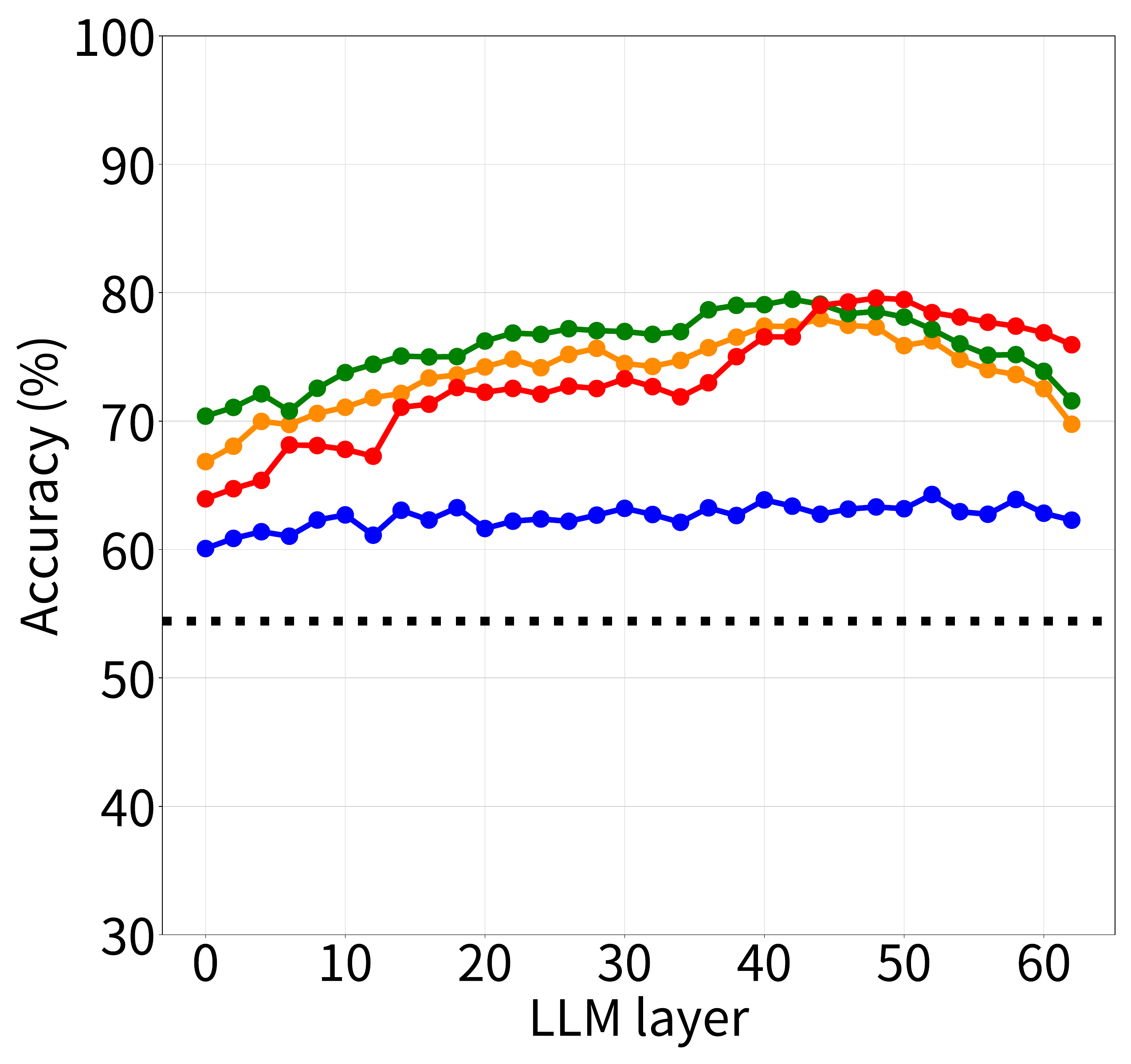}
    {\small \hspace*{1.8em}Word recognition}
  \end{minipage}
  \hfill
  \begin{minipage}{0.245\linewidth}
    \centering
    \includegraphics[height=0.96\linewidth,clip]{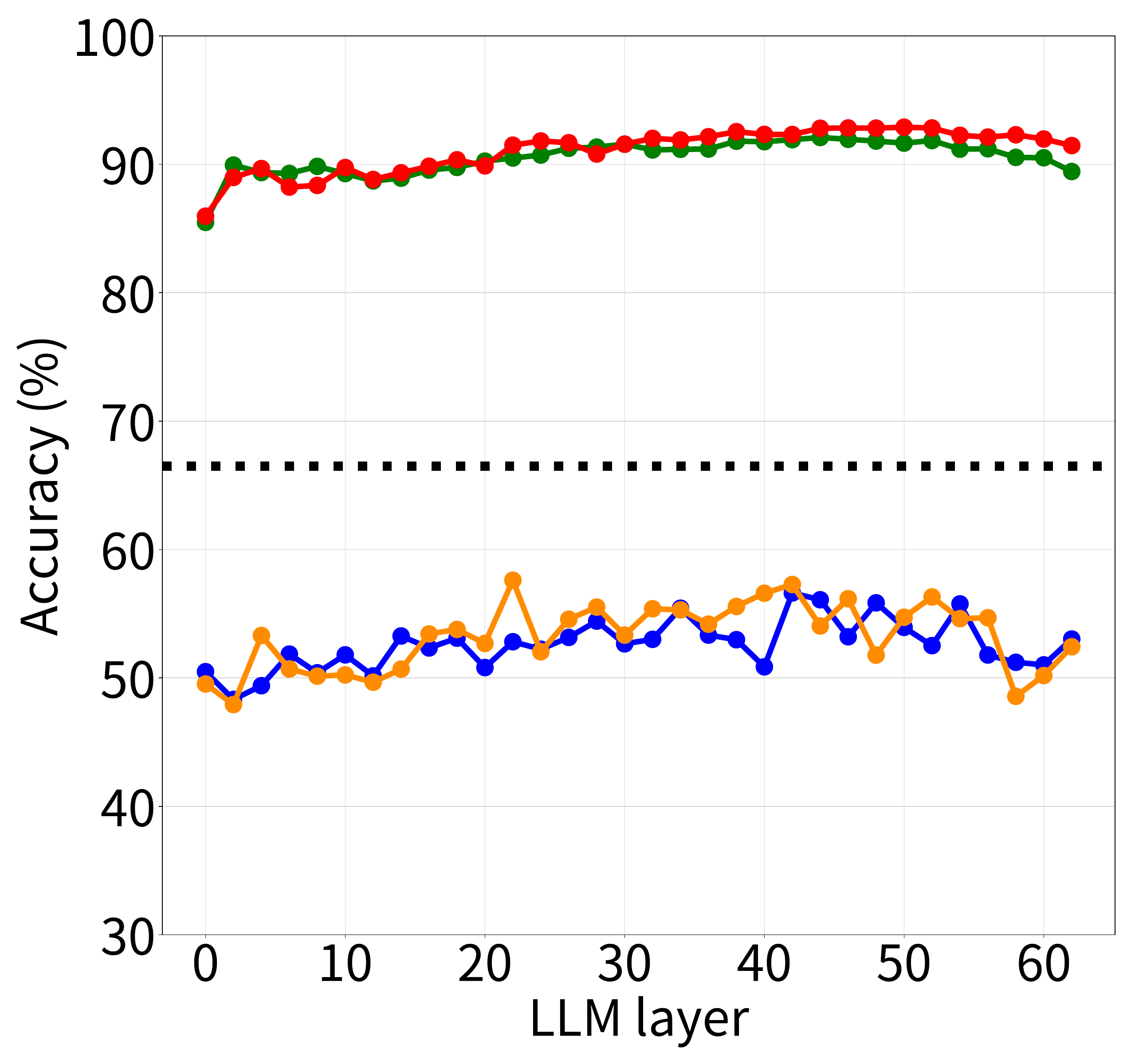}
    {\small \hspace*{1.8em}Structure understanding}
  \end{minipage}
  \hfill
  \begin{minipage}{0.245\linewidth}
    \centering
    \includegraphics[height=0.96\linewidth,clip]{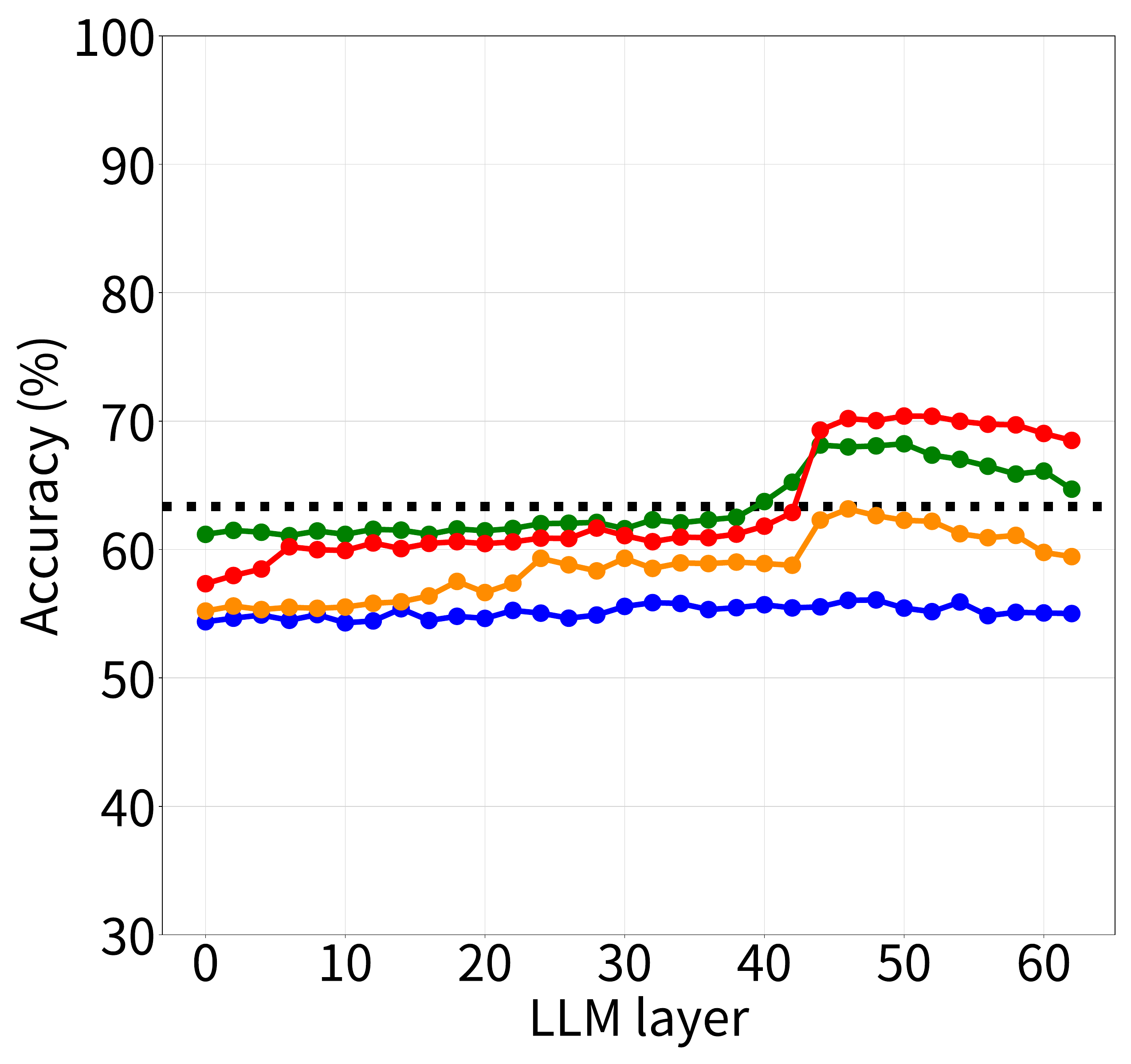}
    {\small \hspace*{1.8em}Figure understanding}
  \end{minipage}
  \par\vspace{0.8em}
  {\small (a) Linear probing results of \textbf{Qwen2.5-VL 32B.}
  }
\end{minipage}

\vspace{1em}

\begin{minipage}{\textwidth}
\centering  
  \begin{minipage}{0.245\linewidth}
    \centering
    \includegraphics[height=0.96\linewidth,clip]{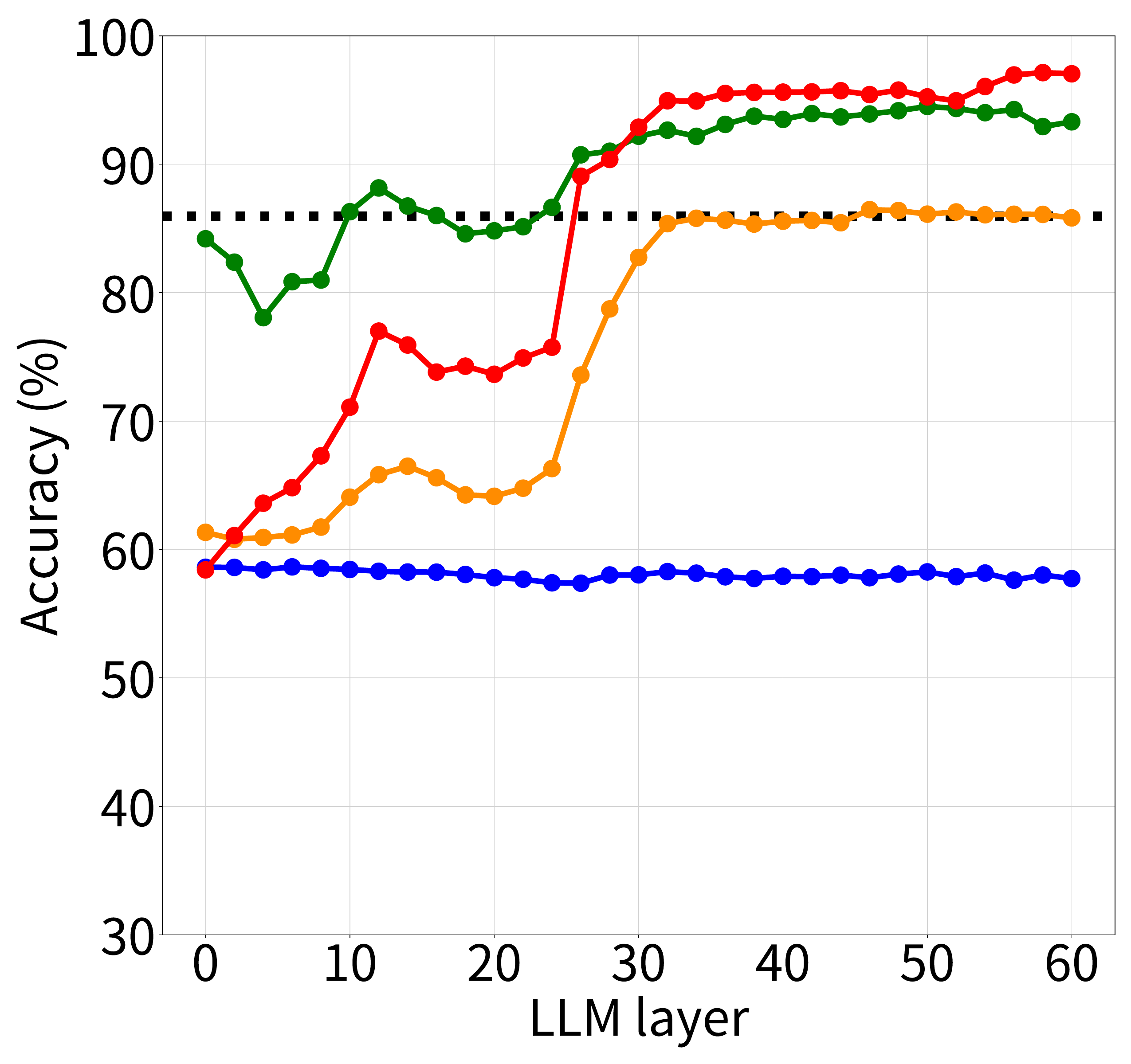}
    {\small \hspace*{1.8em}Visual attributes recognition}
  \end{minipage}
  \hfill
  \begin{minipage}{0.245\linewidth}
    \centering
    \includegraphics[height=0.96\linewidth,clip]{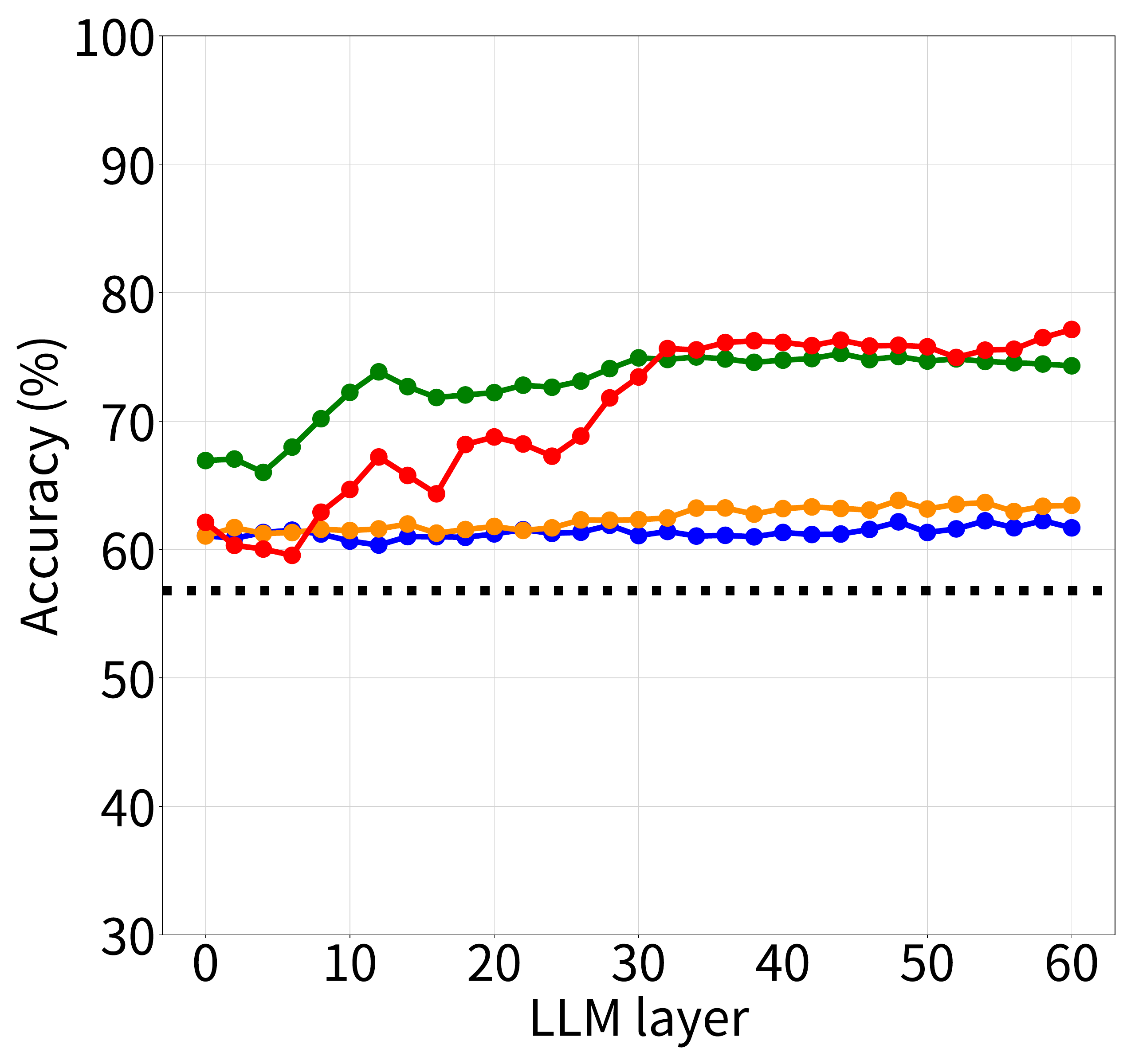}
    {\small \hspace*{1.8em}Word recognition}
  \end{minipage}
  \hfill
  \begin{minipage}{0.245\linewidth}
    \centering
    \includegraphics[height=0.96\linewidth,clip]{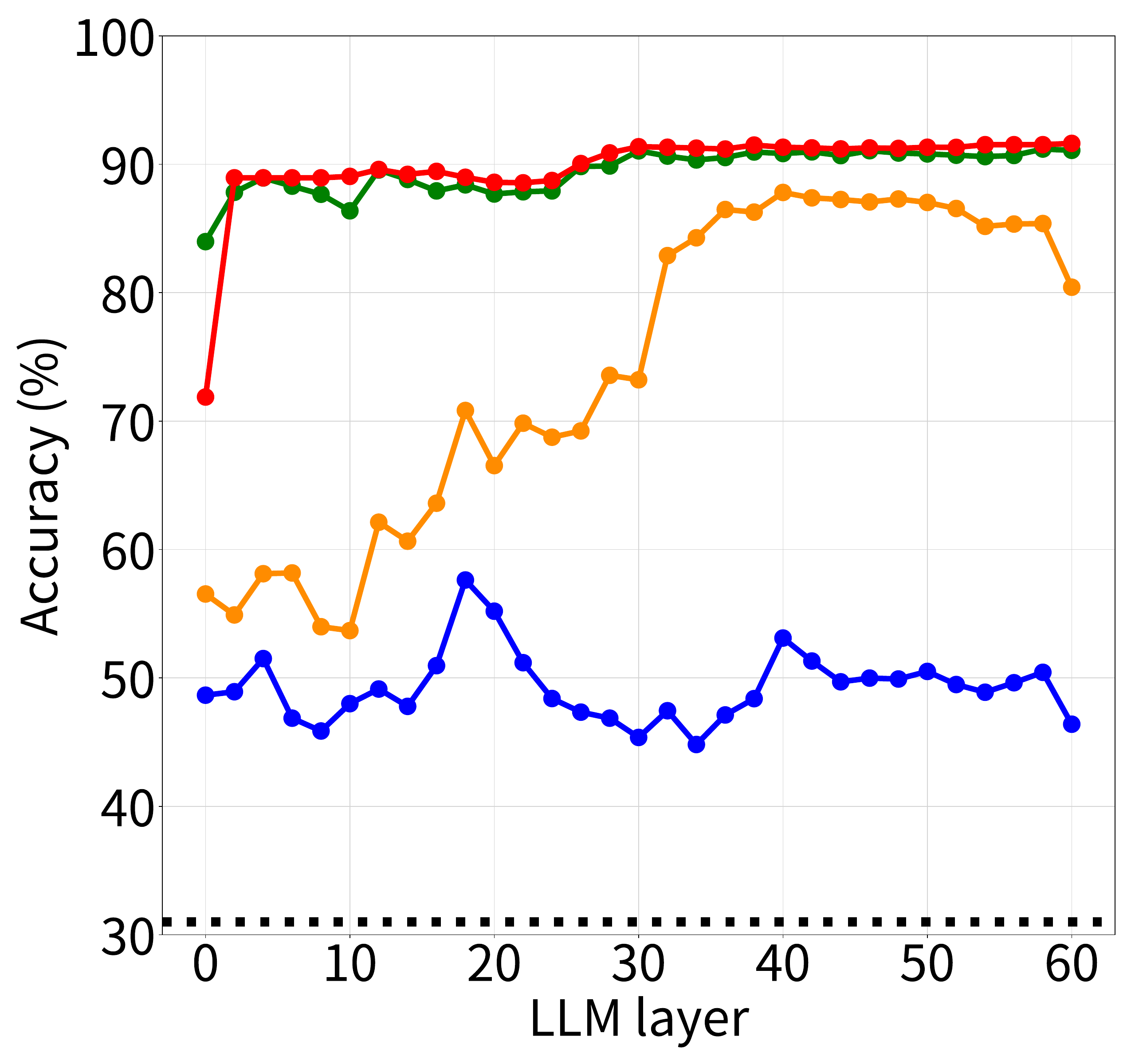}
    {\small \hspace*{1.8em}Structure understanding}
  \end{minipage}
  \hfill
  \begin{minipage}{0.245\linewidth}
    \centering
    \includegraphics[height=0.96\linewidth,clip]{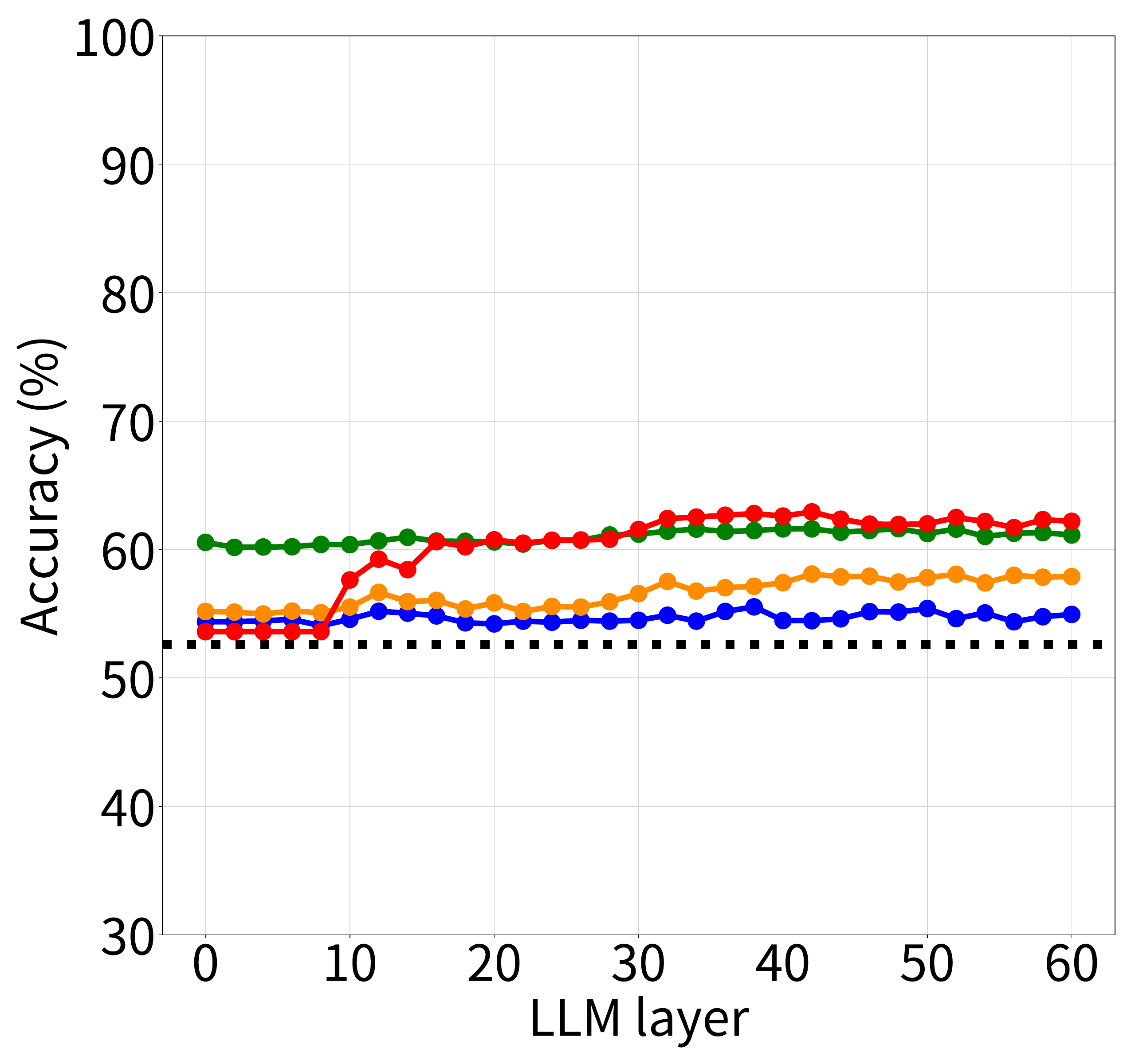}
    {\small \hspace*{1.8em}Figure understanding}
  \end{minipage}
  \par\vspace{0.8em}
  {\small (b) Linear probing results of \textbf{Gemma3 27B.}}
\end{minipage}

\vspace{1em}

\begin{minipage}{\textwidth}
\centering  
  \begin{minipage}{0.245\linewidth}
    \centering
    \includegraphics[height=0.96\linewidth,clip]{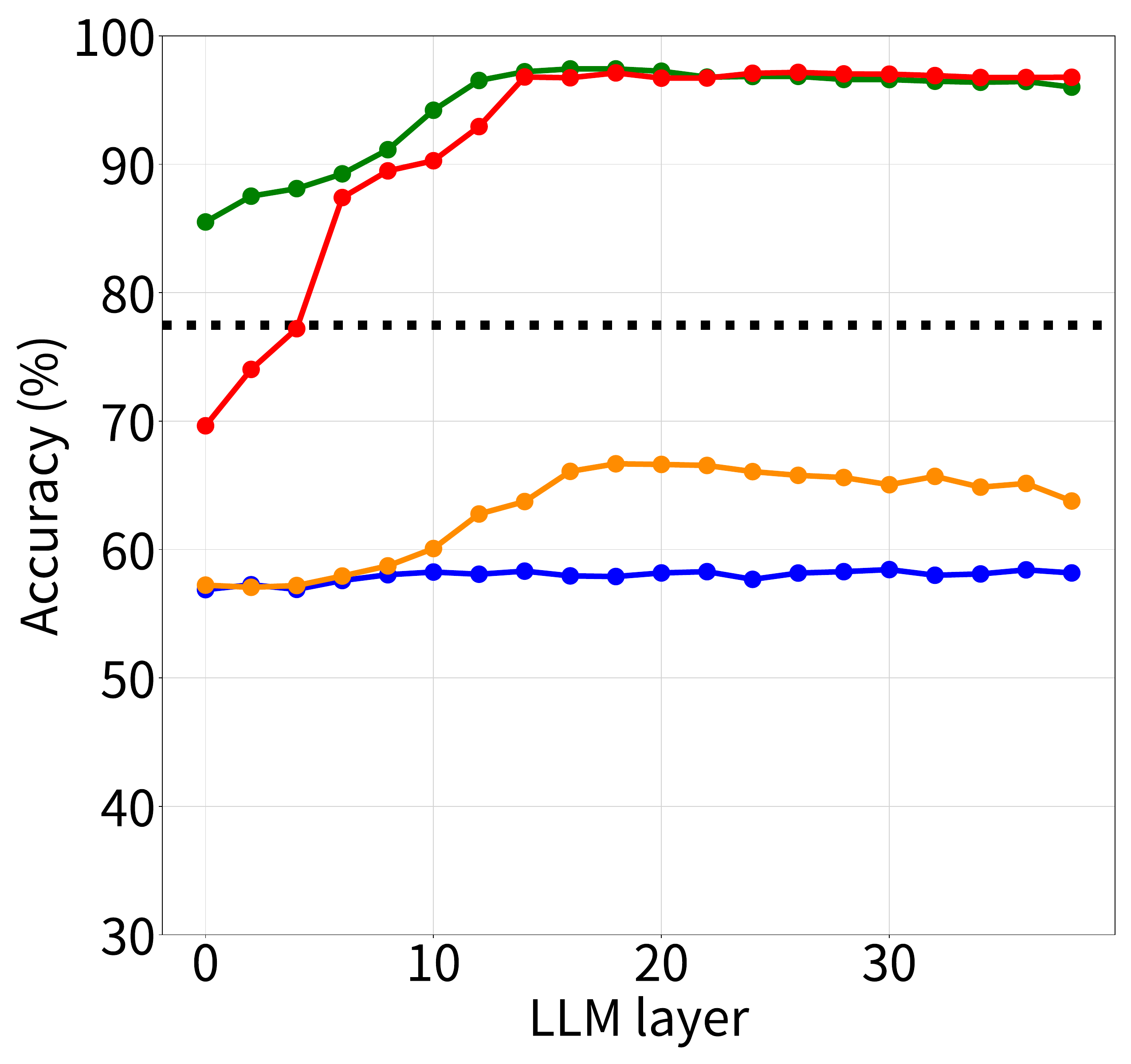}
    \par{\small \hspace*{1.8em}Visual attributes recognition}
  \end{minipage}
  \hfill
  \begin{minipage}{0.245\linewidth}
    \centering
    \includegraphics[height=0.96\linewidth,clip]{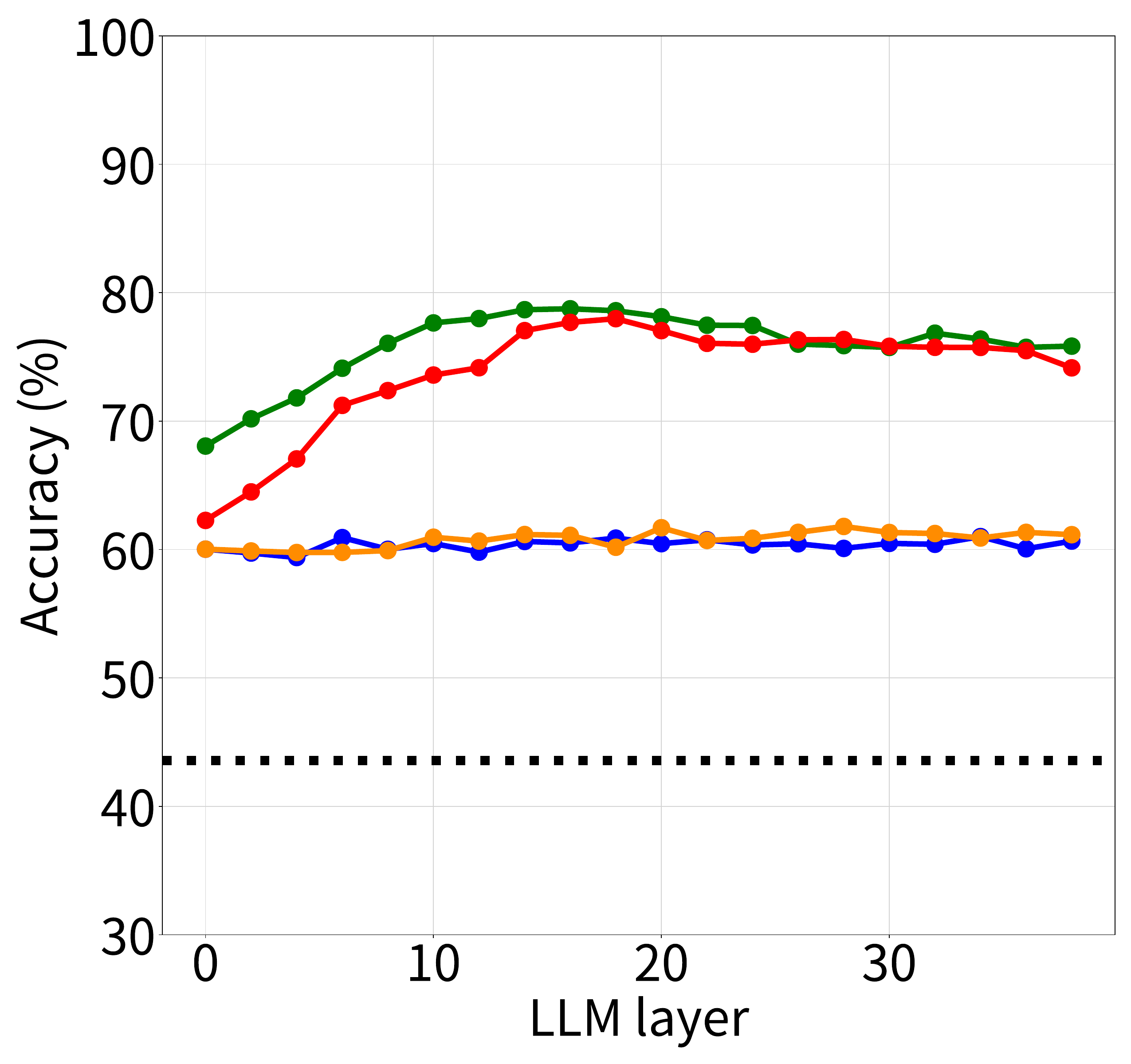}
    \par{\small \hspace*{1.8em}Word recognition}
  \end{minipage}
  \hfill
  \begin{minipage}{0.245\linewidth}
    \centering
    \includegraphics[height=0.96\linewidth,clip]{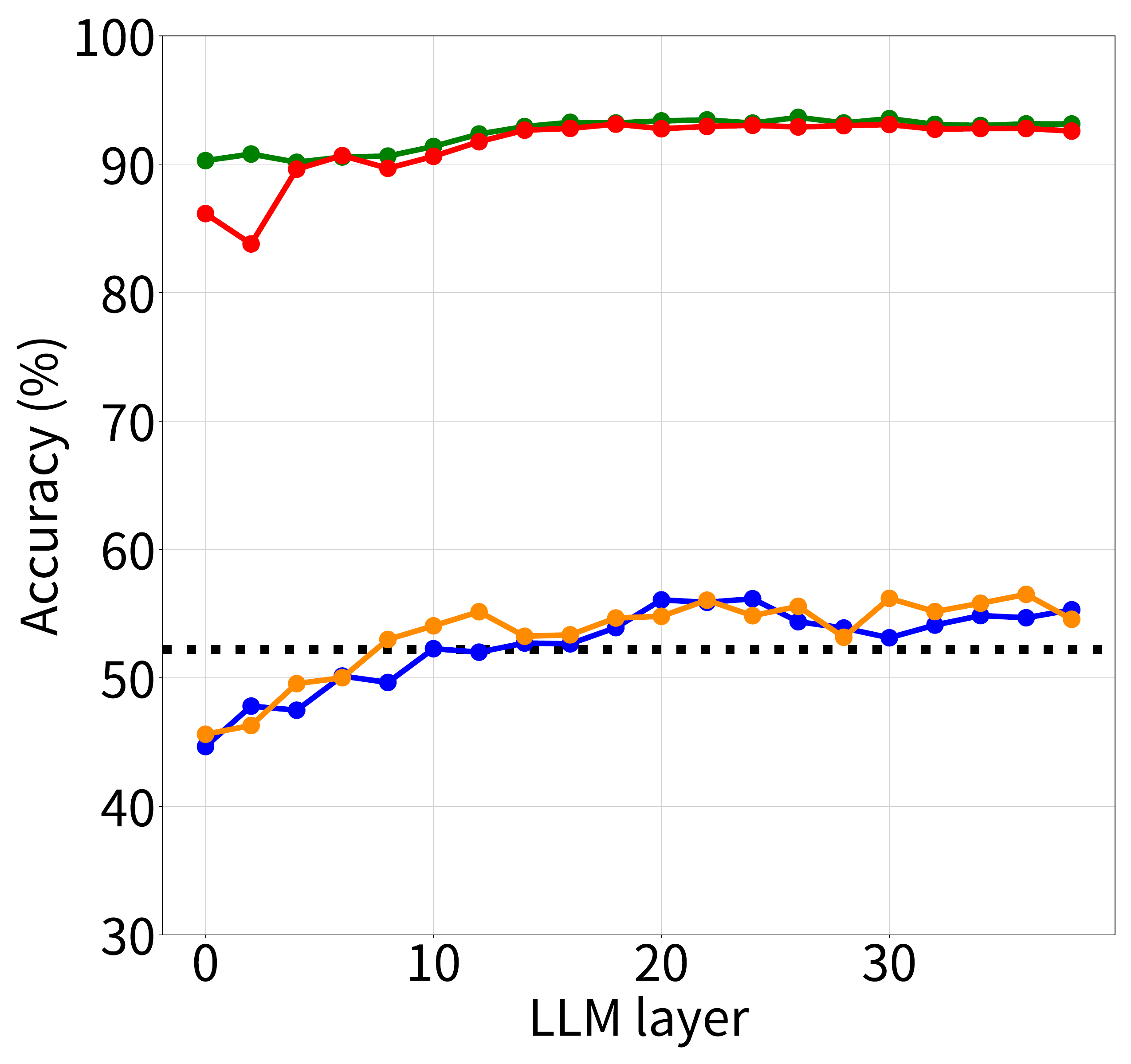}
    \par{\small \hspace*{1.8em}Structure understanding}
  \end{minipage}
  \hfill
  \begin{minipage}{0.245\linewidth}
    \centering
    \includegraphics[height=0.96\linewidth,clip]{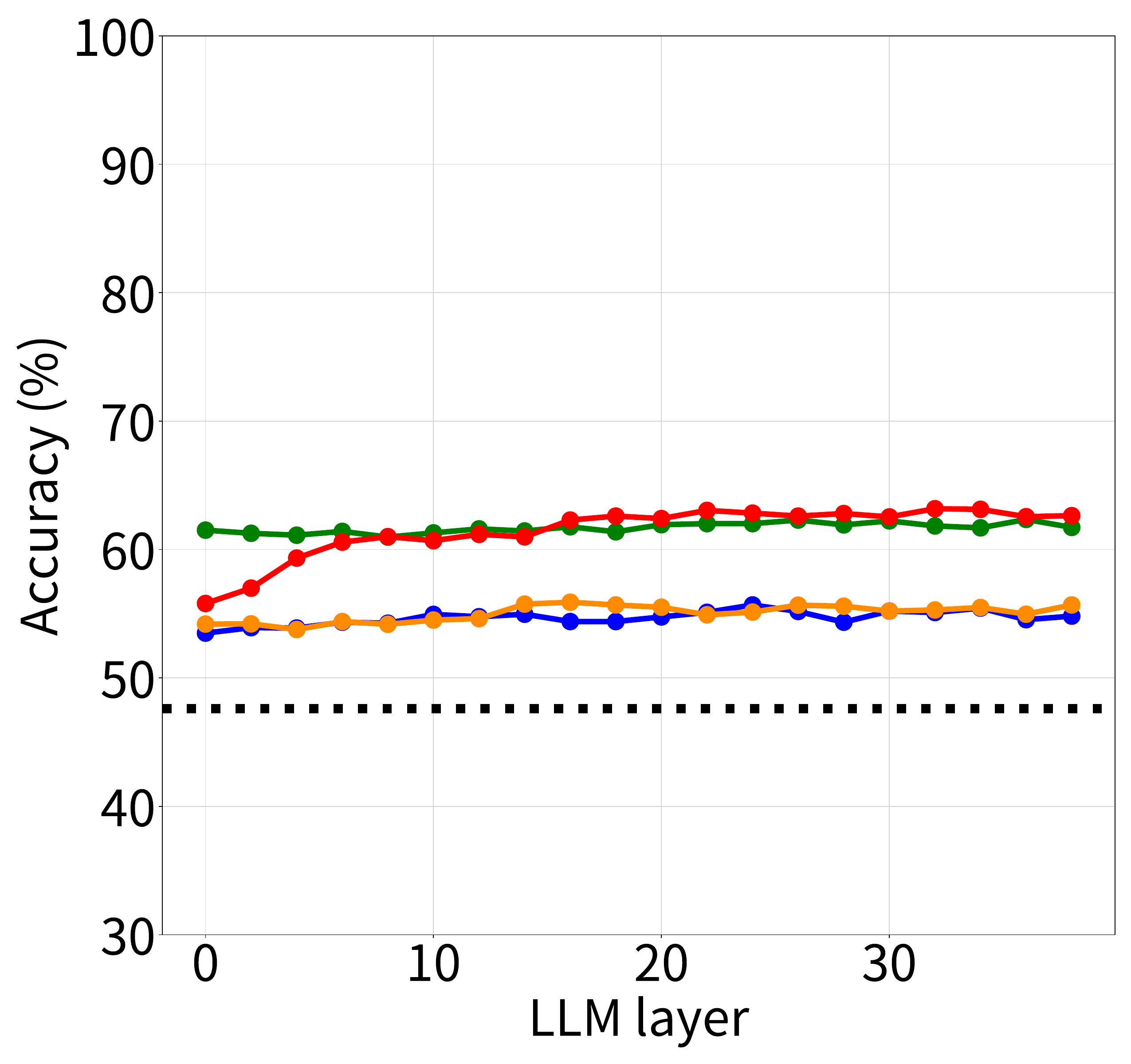}
    \par{\small \hspace*{1.8em}Figure understanding}
  \end{minipage}
  \par\vspace{0.8em}
  {\small (c) Linear probing results of \textbf{LLaVA-NeXT 13B.}}
\end{minipage}
\caption{\textbf{Linear probing accuracy} at each layer (line plot) and text-response accuracy (horizontal black dotted line).
Token types used in linear probing are divided into four categories: \hlc[nblue!30]{image-token}, \hlc[ngreen!30]{text-token}, \hlc[oorange!30]{all-token}, and \hlc[nred!30]{last-token}.
The vertical axis represents accuracy, and the horizontal axis corresponds to the LLM layers, plotted every two layers.}
\label{fig:linear_probe}
\end{figure*}

\subsection{Evaluation metrics}
\label{eval_metrics}
\paragraph{Linear probing accuracy.}
We report the classification accuracy for each classifier. 
The linear probing accuracy is computed as
\[
\mathrm{A}_{\text{LP}} =
\frac{1}{N}
\sum_{i=1}^{N}
\mathbf{1}\!\left(
\arg\max_{y \in \{0,1\}} z_{i,y} = y_i
\right),
\]
where $z_{i,y}$ denotes the output logits of the linear classifier,
$y_i$ is the ground-truth label, $N$ is the number of evaluation samples, and $\mathbf{1}(\cdot)$ is the indicator
function. Since all tasks are balanced binary classification problems, the
chance level is 50\%.

\paragraph{Response accuracy.}
We instruct the LVLMs to answer the same binary tasks as linear probing using the template
“\textsf{\textlangle question\textrangle\hspace{0.1cm}If yes, answer 1; if no, answer 0. Please answer with numbers only.}”
Here, \textsf{\textlangle question\textrangle} is a placeholder for the query from each sample.
We then extract the predicted label by checking whether the generated response contains the corresponding symbol (0, 1, yes, or no).
Response accuracy is computed as the proportion of samples where this extracted label matches the ground‑truth label.
The response accuracy is
computed as
\[
\mathrm{A}_{\text{resp}} =
\frac{1}{N}
\sum_{i=1}^{N}
\mathbf{1}(\tilde{y}_i = y_i),
\]
where $\tilde{y}_i$ denotes the label extracted from the generated response,
$y_i$ is the ground-truth label, $N$ is the number of evaluation samples, and
$\mathbf{1}(\cdot)$ is the indicator function.

\paragraph{Gap between internal representations and responses.}
We define the gap between internal representations and responses as the difference between the maximum linear probing accuracy across layers and the response accuracy.
Formally,
\[
\mathrm{Gap} = \max_{l \in \mathcal{L}} A_{\text{LP}}(l) - A_{\text{resp}},
\]
where $\mathcal{L}$ denotes the set of LLM layers, $A_{\text{LP}}(l)$ is the linear probing accuracy obtained from layer $l$, and $A_{\text{resp}}$ is the response accuracy of the LVLM on the same task.
This comparison allows us to evaluate whether information encoded in internal representations is actually utilized during response generation.

\subsection{Implementation Details}

We initialize the linear classifiers with Xavier uniform initialization and zero bias.
We train the linear classifiers using the Adam~\cite{kingma2015adam} optimizer with an initial learning rate of 1e-3. 
The learning rate is scheduled using a cosine annealing schedule during training. 
All experiments are conducted on a single A100-80G GPU. 
The batch size is set to 256, and the model is trained for one epoch.
For each task, we train separate classifiers on the representations extracted from each layer and each token type.
Experiments are conducted using Qwen2.5-VL 32B Instruct~\cite{Qwen2.5-VL}, Gemma3 27B IT~\cite{team2025gemma} and LLaVA-NeXT 13B~\cite{liu2024llavanext}.

% =====================================================================
\section{Analysis of Linear Probing Results} 
\label{sec:internal_mechanism}

% ------------------------------
\subsection{Gap of Internal Representations and Responses}

\paragraph{Does a gap between internal representations and responses exist?}
As shown in Figure~\ref{fig:linear_probe}, we found that the highest linear probing accuracy exceeds response accuracy across all models and tasks.
In many cases, the highest linear probing accuracy at each layer surpasses the response accuracy even in lower layers.
This observation reveals that the models cannot fully express their internal representations in their responses.

% ------------------------------
\subsection{Further Discussion on Internal Representations}
\label{subsec:further_discussion}

\paragraph{Which token types encode the most information about the answer?}
As shown in Figure~\ref{fig:linear_probe}, across many tasks and models, the last token exhibits the highest linear probing accuracy, indicating that the last token provides a more linearly encoded representation of the ground-truth label than other tokens.

To examine this phenomenon in more detail, we discuss it by token type.
Across most models and tasks, the linear probing accuracy on the \hlc[nblue!30]{image-token} remains close to the chance level of 50\%. 
One possible explanation is that the images are provided before the question.
Another possible explanation is the mean-pooling applied to image tokens, which may dilute localized visual information.
Nevertheless, the results consistently indicate that image tokens encode weaker linearly separable information required to solve the question compared to text tokens and all tokens.

The \hlc[ngreen!30]{text-token} exhibits high linear probing accuracy. 
Although text tokens often achieve higher linear probing accuracy than other token types in the early layers, their accuracy frequently decreases near the final layers. 
This behavior suggests that, in the initial layers, information from the question is primarily preserved within the text tokens, leading to high performance. 
However, accuracy decreases near the final layers, possibly because many text-token representations contain information irrelevant to the answer.

For the \hlc[oorange!30]{all-token} setting, linear probing accuracy exhibits much greater variability, with some tasks and models achieving high accuracy and others markedly lower accuracy. 
This variability is likely due to the heterogeneous information carried by different token types.
When these tokens are mean-pooled, the mixture of disparate information may obscure representations relevant to the information required to solve the question, making accurate classification more difficult.

For the \hlc[nred!30]{last-token} setting, many models and tasks exhibit low linear probing accuracy in the early layers, followed by a substantial increase in accuracy in the higher layers. 
This pattern likely arises because the last token is most directly involved in generation and therefore tends to accumulate the information necessary to produce the answer as the computation progresses through the later layers.

These observations suggest that the last token is the most suitable for linear probing, as it contains the strongest linearly encoded information required to solve the question.

\paragraph{How does the information required to solve the question become linearly encoded across layers?}
We analyze this question using the last token, as it most linearly encodes information required to solve the question, as indicated in the first question of Section~\ref{subsec:further_discussion}.
As shown by the red line plot in Figure~\ref{fig:linear_probe}, the last token linear probing accuracy tends to be higher in the upper layers than in the lower layers. 

A more detailed inspection reveals several sharp increases in linear probing accuracy across layers.
These abrupt improvements appear at approximately the same layers across tasks within each model. 
This suggests that the layer in which information changes within the model remains the same across tasks.
Prior works~\cite {Zhang_2025_CVPR,Jiang_2025_CVPR,jain2025visper_lm,Bi_2025_CVPR,zhang2025shallow,li2025causal} suggest that LLMs in LVLMs primarily perform general visual processing in the lower layers, process visual information required to solve the question in the middle layers, and primarily engage in linguistic processing in upper layers.
Based on prior works and our observations, these sharp increases are likely associated with a shift in the type of processing occurring in the model.

Although this trend is consistent across models, the magnitude of these increases varies by task. 
Tasks that show larger increases generally involve object-level recognition (visual attributes recognition and word recognition), whereas tasks with smaller increases typically require relational reasoning (structure understanding and figure understanding). 
This observation suggests that object-level information can be transformed into representations that are more easily captured by a linear probe, while information required for relational reasoning is harder to encode in a linearly separable form.
Based on prior works mentioned above and our observations, our results suggest that current models still have limitations in performing reasoning and linguistic processing.

Notably, the layer with the highest accuracy is often not the final layer. 
This suggests that representations in which answer information is most linearly encoded are located in intermediate layers rather than the final one.
Consistent with this observation, prior work has shown that intermediate layers often provide more informative representations than final layers~\cite{skean2025layer,wu2026hidivdrop}.
One possible explanation for the decline after the peak is that representations may be optimized for next-token prediction rather than for preserving linearly separable signals of the correct answer as computation proceeds toward the final layers.
The model may perform additional reasoning steps after the answer signal emerges, which can modify or dilute the earlier representation. 
These factors suggest that the gap between internal representations and responses may partly arise from the dynamics of autoregressive generation.

\begin{tcolorbox}[
    boxrule=0.2mm, 
    boxsep=0mm, 
    colback=white,
]
\textit{\textbf{Takeaway messages in Section~\ref{sec:internal_mechanism}:}} 
\begin{itemize}
\item Internal representations of LVLMs are not fully reflected in their responses, even for VDU tasks.

\item Information required to solve the question is often more accessible in intermediate layers than in the final layer.
\end{itemize}
\end{tcolorbox}

% =====================================================================
\section{Experiments for Bridging the Gap}
\label{sec:tuning}

As noted in the takeaways of Section~\ref{sec:internal_mechanism}, we found that LVLMs suffer from the limitation that internal representations are not always accurately expressed in responses for VDU, and linear probing accuracy often peaks not at the final layer but at intermediate layers. 
These findings suggest that bridging the gap between internal representations and responses may require a method targeting the intermediate layers.
To test this hypothesis, we examine whether fine-tuning selected intermediate layers can bridge this gap.
In contrast to prior methods that fine-tune only specific intermediate layers in LLMs~\cite{liu-etal-2024-aflora,NEURIPS2024_68716328,guangyuanunderstanding,yao2024layer,electronics13112140,app151910434,hui-etal-2025-hft,wei2025flexora}, LVLMs~\cite{reinhardt2024improving,wang-etal-2025-activating}, and others~\cite{devoto2024adaptive,li2025enhancing}, we select fine-tuning layers based on insights derived from linear probing analysis of LVLMs for VDU.

\begin{table*}[t] 
    \centering
    \small
    \caption{\textbf{Response accuracy and linear probing accuracy and the gap between them of models fine-tuned intermediate layers on classification tasks.}
\textbf{Bold} values indicate the best for each task, and \underline{underlined} values indicate the second-best.
($\cdot$) represents the effective amount of trainable parameters, normalized such that full-layer tuning for the entire training corresponds to 100.
The value reflects both the number of unfrozen parameters and the number of epochs for which they are trained.
Accuracy is reported in percentages.
    }
    \label{tab:acc_and_gap}
    \begin{tabular}{l|cc|ccccccccc}  
        \toprule 
           Metrics  & Base & All & Lower & Middle & Upper & L--M & M--U & L$\rightarrow$M & M$\rightarrow$L & M$\rightarrow$U & U$\rightarrow$M \\
          & (0) & (100) & (20)  &  (48)  &  (31)  & (69) & (80) & (35) & (35) & (40) & (40)\\ 
         \midrule
         \multicolumn{12}{c}{\textit{Visual attributes recognition}} \\
         Response accuracy ($\uparrow$) & 67.96 & 92.29 & 65.82 & \underline{93.17} & 68.60 & 90.48 & 91.43 & 76.55 & \textbf{95.62} & 89.09 & 88.49 \\ 
         Linear probing accuracy ($\uparrow$) & 92.78 & 96.60 & 93.24 & 96.39 & 93.12 & 96.16 & 96.65 & 96.10 & \underline{98.12} & 94.70 & \textbf{98.23}  \\
         Gap ($\downarrow$) & 24.82 & 4.31  & 27.42  & \underline{3.22} & 24.52 & 5.68 & 5.22 & 19.55 & \textbf{2.50} & 5.61 & 9.74\\
         \multicolumn{12}{c}{\textit{Word recognition}} \\
          Response accuracy ($\uparrow$) & 54.43 & 68.06 & 58.33 & 70.00 & 54.42 & \textbf{74.66} & 63.04 & 70.98 & \underline{73.33} & 65.88 & 64.84 \\ 
          Linear probing accuracy ($\uparrow$) & 79.84 & 84.39 & 82.24 & 84.40 & 79.89 & 84.61 & 83.72 & \underline{86.05}  & 85.95 & 82.61 & \textbf{86.10} \\
         Gap ($\downarrow$) & 25.41 & 16.33 &23.91 & 14.40 & 25.47 & \textbf{9.95} & 20.68 & 15.07 & \underline{12.62} & 16.73 & 21.26\\
         \multicolumn{12}{c}{\textit{Structure understanding}} \\
          Response accuracy ($\uparrow$)& 66.50 & 83.42 & 65.94 & 82.87 & 67.70 & \underline{84.99} & 82.23 & \textbf{88.31} & 82.08 & 83.24 & 68.55   \\ 
          Linear probing accuracy ($\uparrow$) & 92.89 & 92.94 &  92.93 & 93.15 & 93.05 & 92.96 & 93.19 & \underline{93.29} & 93.09 & 93.15 & \textbf{93.38}\\
         Gap ($\downarrow$) & 26.39 & 9.52 & 26.99 & 10.28 & 25.35 & \underline{7.97} & 10.96 & \textbf{4.98} & 11.01 & 9.91 & 24.83 \\
         \multicolumn{12}{c}{\textit{Figure understanding}} \\
          Response accuracy ($\uparrow$) & 63.34 & 66.54 & 65.49 & 66.36 & 63.30 & 66.72 & 65.96 &  \textbf{70.33} & 67.94 & 65.95 & \underline{68.45} \\
          Linear probing accuracy ($\uparrow$) & 70.48 & 72.15 & 72.38 & 72.07 & 70.33 & 72.64 & 71.49 & \textbf{74.52} & \underline{73.95} & 71.66 & 73.64 \\
         Gap ($\downarrow$)  & 7.14 & 5.61 & 6.89 & 5.71 & 7.03 & 5.92 & 5.53 & \textbf{4.19} & 6.01 & 5.71 & \underline{5.19} \\
        \bottomrule 
\end{tabular}
\end{table*}

\subsection{Experimental Setup}
\paragraph{Model and layer configuration.}
We use Qwen2.5-VL 32B Instruct as the base model. 
We divide the LLM layers according to where linear probing accuracy shows major increases: layers 0-12 (up to the first rise) as the lower layers, layers 13-43 (up to the second rise) as the middle layers, and layers 44-63 as the upper layers.
We denote these groups as “Lower”, “Middle”, and “Upper”.
We then conduct experiments by fine‑tuning different combinations of these layer groups in both single‑step and two‑step settings.

\paragraph{Fine‑tuning strategies.}
Our fine-tuning strategies consist of two settings: \textbf{single-step} and \textbf{two-step}. In the \textbf{single‑step} setting, we fine‑tune either a single group or two adjacent groups of layers in a single training run.
We denote the range of fine-tuned layers using labels such as “Lower” or “L--M”.
For example, “L--M” indicates that layers 0-43 (i.e., the Lower and Middle groups) are fine-tuned in a single step.
“All” denotes that all LLM layers are fine-tuned in a single step. In the \textbf{two‑step} setting, the layers fine-tuned in the first and second steps do not overlap, and each step is trained for the same number of epochs.
We denote this as “L→M”, meaning that the Lower layers are fine‑tuned in the first step and the Middle layers in the second.

We ensure that all model variants for a given task are trained for the same total number of epochs.
In all configurations, the vision encoder and the projector are frozen, and the lm\_head is unfrozen.

\paragraph{Evaluation tasks and metrics.}
We conduct experiments on both the classification tasks (defined as linear probing tasks in Section~\ref{subsec:data}) and on document VQA tasks.
For document VQA tasks, we use the DocVQA~\cite{mathew2021docvqa} and InfographicVQA~\cite{mathew2022infographicvqa}.
For the classification tasks, we evaluate response accuracy and linear probing accuracy and the gap between them, while in document VQA tasks, we use the Average Normalized Levenshtein Similarity (ANLS)~\cite{Biten_2019_ICCV} to measure textual response quality.

Note that because document VQA tasks require open-ended text generation, linear probing cannot be applied.

\paragraph{Implementation details.}
We perform full fine‑tuning on eight A100-80G GPUs in all settings.
For classification tasks, we use a learning rate of 1e‑6 and a batch size of 2048 and train each model for 10 epochs.
In two-step configurations, each step is trained for 5 epochs.
For the document VQA tasks, based on prior work~\cite{Liu_2024_CVPR}, we use a learning rate of 1e‑5 and a batch size of 128 and train each model for 2 epochs.
The model is trained only on tasks that match the evaluation task.
We perform training in bfloat16 precision with DeepSpeed~\cite{rajbhandari2020zero} (ZeRO Stage 3) and FlashAttention-2~\cite{dao2023flashattention2} enabled.
The linear probing configuration is identical to that described in Section~\ref{sec:lp}.

\subsection{Results and Discussion}

\begin{table*}[t] 
    \centering
    \small
    \caption{
    \textbf{Response quality and training time of models fine-tuned intermediate layers on document VQA tasks.}
ANLS and total training time (minutes and seconds) are reported. 
For two-step training, training time values represent the sum of the first step (up to the corresponding checkpoint) and the second step. 
($\cdot$) represents the effective amount of trainable parameters, normalized such that full-layer tuning for the entire training corresponds to 100.
The value reflects both the number of unfrozen parameters and the number of epochs for which they are trained.
Because document VQA tasks require open-ended text generation, linear probing cannot be applied.}
    \label{tab:vqa_tuning}
    \begin{tabular}{l|cc|ccccccccc} 
        \toprule 
        Metrics  & Base & All & Lower & Middle & Upper & L--M & M--U & L$\rightarrow$M & M$\rightarrow$L & M$\rightarrow$U & U$\rightarrow$M \\
        &  (0) & (100) & (20)  &  (48)  &  (31)  & (69) & (80) & (35) & (35) & (40) & (40)\\
        \midrule 
        \multicolumn{12}{c}{\textit{DocVQA}} \\
        ANLS ($\uparrow$) & 0.9380 & 0.9474 & 0.9379 & 0.9451 & 0.9356 & 0.9430 & \textbf{0.9492} & 0.9418 & 0.9386 & \underline{0.9477} & 0.9410   \\
        Training time & 00:00 & 60:44 & 30:16 &  34:52 & 31:49 & 37:42 &  38:25 &  30:14 
        & 29:42 
        &  30:16 
        & 30:51 
        \\
        \multicolumn{12}{c}{\textit{InfographicVQA}} \\
         ANLS ($\uparrow$) & 0.8352 & 0.8337 & 0.8320 & 0.8341 & 0.8299 & 0.8298 & \textbf{0.8393} & 0.8295 & 0.8306 & \underline{0.8349} & 0.8306\\
         Training time & 00:00 & 40:31 & 20:57 & 23:27 & 21:30  & 26:56 & 25:43 & 19:47 
         & 20:39 
         & 19:44 
         & 20:17 
         \\
        \bottomrule 
    \end{tabular}
\end{table*}

\paragraph{Does all-layer fine-tuning close the gap between internal representations and responses?}
Table~\ref{tab:acc_and_gap} shows response accuracy and the gap between the highest linear probing accuracy from the last-token and response accuracy.
In all cases, linear probing accuracy exceeds response accuracy.
Focusing on the results of the all-layer fine-tuned model (All), we observe that response accuracy consistently improves after fine‑tuning. 
However, the gap is not fully closed.
These results suggest that the gap between linear probing accuracy and response accuracy cannot be explained solely by the model's lack of all-layer fine-tuning.

\paragraph{Is fine-tuning intermediate layers effective for reducing the gap?}

As shown in Table~\ref{tab:acc_and_gap}, we observe that fine-tuning intermediate layers consistently outperform all‑layer fine‑tuning in response accuracy and linear probing accuracy, and narrow the gap between internal representations and responses.
We also observed that the trends for models with high response accuracy and those with small gaps are consistent.

\paragraph{What is the relationship between tasks and important groups of layers?}

To examine response quality trends of intermediate layer fine-tuning in a more practical setting, we also evaluate the model on the document VQA tasks (Table~\ref{tab:vqa_tuning}).
As shown in Table~\ref{tab:acc_and_gap} and Table~\ref{tab:vqa_tuning}, all of the best‑performing and the second-performing models include middle layers within the fine-tuned layer groups.
This indicates that middle layers, which capture visual features directly relevant to answering the question, serve as a key target for reducing the gap.
In classification tasks, fine-tuning either the middle layers or the combined L-M layers yields the highest accuracy for the single‑step setting.
For two-step object‑level recognition tasks, M→L yields better performance, suggesting that enhancing the ability to extract answer‑relevant visual cues before strengthening global visual understanding may facilitate more accurate object comprehension.
For two-step relational reasoning tasks, L→M performs better, suggesting that first enhancing global understanding and then refining the extraction of answer‑relevant information is more effective.
Thus, the importance of the lower-to-middle layers likely reflects the characteristics of the tasks, which rely more heavily on visual processing than on linguistic reasoning.
On the other hand, in document VQA tasks, performance improves when the upper layers are included along with the middle layers.
This is likely because the tasks require linguistic reasoning.

\paragraph{How efficient is fine-tuning of selected intermediate layers?}

As shown in Table~\ref{tab:vqa_tuning}, for the M-U configuration, which achieved the greatest improvement in DocVQA, performance gains were obtained while reducing the number of trainable parameters by approximately 20\% and the training time by approximately 35\%.
Fine-tuning selected layer not only improves both linear probing accuracy and response accuracy, and narrows the gap between internal representations and responses, but also enhances training parameter and time efficiency.

\begin{tcolorbox}[
    boxrule=0.2mm,  
    boxsep=0mm,    
    colback=white,
]
\textit{\textbf{Takeaway message in Section~\ref{sec:tuning}:}} \\
The gap between internal representations and responses cannot be explained by the lack of all-layer fine-tuning alone.
However, this gap can be mitigated by fine‑tuning the appropriate intermediate layers for solving the task.
\end{tcolorbox}

% =======================================================================
\section{Conclusion}
\label{sec:conclusion}

In this work, we conducted the first detailed investigation of the internal representations of LLMs within LVLMs for VDU using linear probing, addressing the limitation that LVLMs often fail to fully express their internal representations in responses.
Our analysis shows that, even in VDU, LVLMs do not completely convey their internal representations through responses. 
Furthermore, the detailed examination revealed that the information required to solve VDU tasks is most strongly linearly encoded in intermediate layers rather than the final layer. 
Motivated by these findings, we hypothesized that fine-tuning intermediate layers could reduce this gap and conducted an experiment fine-tuning selected intermediate layers. 
Our experiments demonstrate that fine-tuning the most important intermediate layers for answering each task improves both linear probing accuracy and response accuracy while narrowing the gap.
These observations suggest that approaches that leverage intermediate layers hold promise for further reducing the gap between internal representations and responses.
We hope this work encourages further research on understanding and improving the alignment between internal representations and responses in LVLMs.

\paragraph{Limitation.}
In this work, we employ linear probing to analyze the internal representations of LVLMs.
Although linear probing enables systematic comparisons across layers and facilitates quantitative analysis of the relationship between internal representations and model responses, it is less suitable for analyzing more complex information, especially that involved in free-form or complex reasoning.
Addressing this limitation may require developing new probing methods capable of assessing whether complex information is encoded in model representations.
For instance, prior work~\cite{white-etal-2021-non} has explored more expressive probing approaches, such as non-linear structural probes, to capture complex structures encoded in model representations.

{
    \small
    \bibliographystyle{ieeenat_fullname}
    \bibliography{main}
}


\end{document}